\documentclass[10pt,twocolumn,letterpaper]{article}

\usepackage{iccv}
\usepackage{times}
\usepackage{epsfig}
\usepackage{graphicx}
\usepackage{amsmath}
\usepackage{bm}
\usepackage{mathtools}
\usepackage{amssymb}
\usepackage{array}
\usepackage{makecell}
\usepackage{booktabs}
\usepackage{tikz}
\usepackage{pgfplots}
\usepackage{microtype}
\usepackage{caption}
\usepackage{subcaption}
\usepackage{float}

\usepackage[pagebackref=true,breaklinks=true,letterpaper=true,colorlinks,bookmarks=false]{hyperref}
\usepackage[capitalize,noabbrev]{cleveref}

\iccvfinalcopy

\def\iccvPaperID{3495} 

\ificcvfinal\fi

\DeclarePairedDelimiter\abs{\lvert}{\rvert}%
\DeclarePairedDelimiter\norm{\lVert}{\rVert}%

\makeatletter
\let\oldabs\abs
\def\abs{\@ifstar{\oldabs}{\oldabs*}}
\let\oldnorm\norm
\def\norm{\@ifstar{\oldnorm}{\oldnorm*}}
\makeatother

\newcommand\authormark[1]{\textsuperscript#1}

\begin{document}
    \title{Improving Dense Crowd Counting Convolutional Neural Networks\\using Inverse k-Nearest Neighbor Maps and Multiscale Upsampling}

\author{Greg Olmschenk\authormark{1} \hspace{2cm} Hao Tang\authormark{2} \hspace{2cm} Zhigang Zhu\authormark{{1,3}}\\
\authormark{1}The Graduate Center of the City University of New York\\
\authormark{2}Borough of Manhattan Community College - CUNY\\
\authormark{3}The City College of New York - CUNY\\
{\tt\small golmschenk@gradcenter.cuny.edu, htang@bmcc.cuny.edu, zhu@cs.ccny.cuny.edu}
}

\maketitle
    \begin{abstract}
    Gatherings of thousands to millions of people frequently occur for an enormous variety of events, and automated counting of these high-density crowds is useful for safety, management, and measuring significance of an event. In this work, we show that the regularly accepted labeling scheme of crowd density maps for training deep neural networks is less effective than our alternative inverse k-nearest neighbor (i$k$NN) maps, even when used directly in existing state-of-the-art network structures. We also provide a new network architecture MUD-i$k$NN, which uses multi-scale upsampling via transposed convolutions to take full advantage of the provided i$k$NN labeling. This upsampling combined with the i$k$NN maps further improves crowd counting accuracy. Our new network architecture performs favorably in comparison with the state-of-the-art. However, our labeling and upsampling techniques are generally applicable to existing crowd counting architectures.
\end{abstract}

    \section{Introduction}

Every year, gatherings of thousands to millions occur for protests, festivals, pilgrimages, marathons, concerts, and sports events. For any of these events, there are countless reasons to desire to know how many people are present. For those hosting the event, both real-time management and future event planning is dependent on how many people are present, where they are located, and when they are present. For security purposes, knowing how quickly evacuations can be executed and where crowding might pose a threat to individuals is dependent on the size of the crowds. In journalism, crowd sizes are frequently used to measure the significance of an event, and systems which can accurately report on the event size are important for a rigorous evaluation.

Many systems have been proposed for crowd counting purposes, with most recent state-of-the-art methods being based on convolutional neural networks (CNNs). To the best of our knowledge, every CNN-based dense crowd counting approach in recent years relies on using a density map of individuals, primarily with a Gaussian-based distribution of density values centered on individuals labeled in the ground truth images. Often, these density maps are generated with the Gaussian distribution kernel sizes being dependent on a k-Nearest Neighbor ($k$NN) distance to other individuals~\cite{zhang2016single}. In this work, we explain how this generally accepted density map labeling is lacking and how an alternative inverse $k$NN (i$k$NN) labeling scheme, which does not explicitly represent crowd density, provides improved counting accuracy. We will show how a single i$k$NN map provides information similar to the accumulation of many density maps with different Gaussian spreads, in a form which is better suited for neural network training. This labeling provides a significant gradient spatially across the entire label while still providing precise location information of individual pedestrians (with the only exception being exactly overlapping head labelings). We show that by simply replacing density map training in an existing state-of-the-art network with our i$k$NN map training, the testing accuracy of the network improves. This is the first major contribution of the paper.

Additionally, coupling multi-scale upsampling with densely connected convolutional networks~\cite{huang2017densely} and our proposed i$k$NN mapping, we provide a new network structure, MUD-i$k$NN, which performs favorably compared to existing state-of-the-art methods. This network uses multi-scale upsampling with transposed convolutions~\cite{zeiler2010deconvolutional} to make effective use of the full ground truth label, particularly with respect to our i$k$NN labeling scheme. The transposed convolutions are used to spatially upsample intermediate feature maps to the ground truth label map size for comparison. This approach provides several benefits. First, it allows the features of any layer to be used in the full map comparison, where many existing methods require a special network branch for this comparison. Notably, this upsampling, comparison, and following regression module can be used at any point in any CNN, with the only change being the parameters of the transposed convolution. This makes the module useful not only in our specific network structure, but also applicable in future state-of-the-art, general-purpose CNNs. Second, as this allows features which have passed through different levels of convolutions to be compared to the ground truth label map, this intrinsically provides a multi-scale comparison without any dedicated additional network branches, thus preventing redundant parameters which occur in separate branches. Third, because the transposed convolution can provide any amount of upsampling (with the features being used to specify the upsampling transformation), the upsampled size can be the full ground truth label size. In contrast, most existing works used a severely reduced size label map for comparison. These reduced sizes remove potentially useful training information. Although some recent works use full-size labels, they require specially crafted network architectures to accomplish this comparison. Our proposed upsampling structure can easily be added to most networks, including widely used general-purpose networks, such as DenseNet. This proposed network structure is the second major contribution of the paper.

Importantly, these contributions are largely complementary to, rather than alternatives to, existing approaches. Most approaches can easily replace their density label comparison with our proposed i$k$NN map comparison and upsampling map module, with little to no modification of the rest of their method or network architecture. As the i$k$NN label does not sum to the count, the i$k$NN label and map module should go hand-in-hand.

The paper is organized as follows. \cref{sec:relatedwork} discusses related work. \cref{sec:knnmaps} describes the proposed k-nearest neighbor map labeling method and its justification. \cref{sec:knnmaps} proposes our new network architecture for crowd counting, MUD-i$k$NN. \cref{sec:results} presents experimental results on several crowd datasets and analyzes the findings. \cref{sec:conclusions} provides a few concluding remarks.
    \section{Related Work}
\label{sec:relatedwork}
\begin{figure*}
    \centering
    \includegraphics[width=\textwidth]{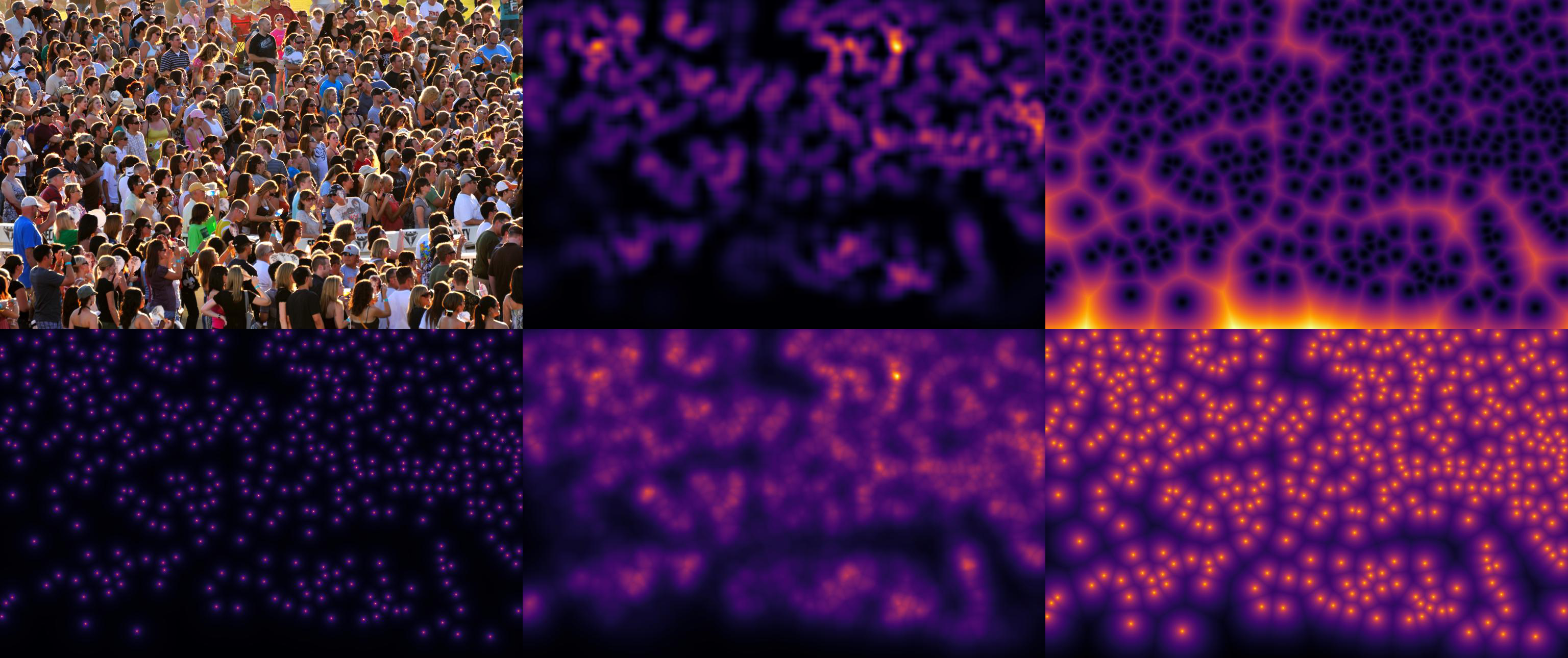}
    \caption{An example of a crowd image and various kinds of labelings. From left to right, on top, there is the original image, the density map, the $k$NN map with $k=1$. On bottom, the inverse $k$NN map with $k=1$, the inverse $k$NN map with $k=3$, and the inverse $k$NN map with $k=1$ shown with a log scaling (for read insight only). Note, in the case of the density map, any values a significant distance from a head labeling are very small. This is in contrast to the inverse $k$NN map, which has a significant gradient even a significant distance from a head position.}
    \label{fig:knnmap}
\end{figure*}
Many works use explicit detection of individuals to count pedestrians~\cite{wu2005detection,lin2010shape,wang2011automatic}. However, as the number of people in a single image increase and a scene becomes crowded, these explicit detection methods become limited by occlusion effects. Early works to solve this problem relied on global regression of the crowd count using low-level features~\cite{chan2008privacy,chen2012feature,chen2013cumulative}. While many of these methods split the image into a grid to perform a global regression on each cell, they still largely ignored detailed spatial information of pedestrian locations.
\cite{lempitsky2010learning} introduced a method of counting objects using density map regression, and this technique was shown to be particularly effective for crowd counting by \cite{zhang2015cross}.
Since then, to the best of our knowledge, every CNN-based crowd counting method in recent years has used density maps as a primary part of their cost function~\cite{idrees2018composition,sam2017switching,sindagi2017cnn,zhang2015cross,zhang2016single,shen2018crowd,li2018csrnet,ranjan2018iterative,shi2018crowd}.

A primary advantage of the density maps is the ability to provide a useful gradient for network training over large portions of the image spatially, which helps the network identify which portion of the image contains information signifying an increase in the count. These density maps are usually modeled by representing each labeled head position with a Dirac delta function, and convolving this function with a 2D Gaussian kernel~\cite{lempitsky2010learning}. This forms a density map where the sum of the total map is equal to the total count of individuals, while the density of a single individual is spread out over several pixels of the map. The Gaussian convolution allows a smoother gradient for the loss function of the CNN to operate over, thereby allowing slightly misplaced densities to result in a lower loss than significantly misplaced densities.

In some works, the spread parameter of the Gaussian kernel is often determined using a $k$-nearest neighbor ($k$NN) distance to other head positions~\cite{zhang2016single}. This provides a form of pseudo-perspective which results in pedestrians which are more distant from the camera (and therefore smaller in the image) having their density spread over a smaller number of density map pixels. While this mapping will often imperfectly map perspective (especially in sparsely crowded images), it works well in practice. Whether adaptively chosen or fixed, the Gaussian kernel size is dependent on arbitrarily chosen parameters, usually fine-tuned for a specific dataset.

In a recent work~\cite{idrees2018composition}, the authors used multiple scales of these $k$NN-based, Gaussian convolved density maps to provide various levels of spatial information, from large Gaussian kernels (allowing for a widespread training gradient) to small Gaussian kernels (allowing for precise localization of density). While this approach effectively integrates information from multiple Gaussian scales, thus providing both widespread and precise training information, the network is left with redundant structures and how the various scales are chosen is fairly ad hoc. Our alternative i$k$NN labeling method supersedes these multiple scale density maps by providing both a smooth training gradient and precise label locations (in the form of steep gradients) in a single label. Our new network structure utilizes a single branch CNN structure for multi-scale regression. Together with the i$k$NN labeling, it provides the benefits of numerous scales of these density maps.

Nearly all these CNN-based approaches use a reduced label size. Some recent works~\cite{shen2018crowd, li2018csrnet} have begun using full resolution labels. In contrast even to these works, we provide a generalized map module which can be added to existing network structures allowing them to take advantage of larger resolutions. Our proposed network is based off the DenseNet201~\cite{huang2017densely}, with our map module added to the end of each DenseBlock. This map module can be added to most CNN architectures with little or no modification to the original architecture.

Our i$k$NN mapping is somewhat related to a distance transform, which has been used for counting in other applications~\cite{arteta2016counting}. However, the distance transform is analogous to a $k$NN map, rather than our i$k$NN. To our knowledge, neither the distance transform nor a method analogous to our i$k$NN labeling has been used for dense crowd counting.

    \section{Inverse $k$-Nearest Neighbor Map Labeling}
\label{sec:knnmaps}
We propose using full image size i$k$NN maps as an alternative labeling scheme from the commonly used density map explained in \cref{sec:relatedwork}. Formally, the commonly used density map~\cite{idrees2018composition,sam2017switching,sindagi2017cnn,zhang2015cross,zhang2016single} is provided by,
\begin{multline}
    D(\bm{x}, f(\cdot)) =\\\sum\limits_{h=1}^H
    \frac{1}{\sqrt{2\pi}f(\sigma_h)}
    \exp{\left(-\frac{
        (x-x_h)^2+(y-y_h)^2
    }{
        2f(\sigma_h)^2
    }\right)}\text{,}
\end{multline}
where $H$ is the total number of head positions for the example image, $\sigma_h$ is a size determined for each head position $(x_h, y_h)$ using the $k$NN distance to other heads positions (a fixed size is also often used), and $f$ is a manually determined function for scaling $\sigma_h$ to provide a Gaussian kernel size. For simplicity, in our work we define $f$ as a simple scalar function given by $f(\sigma_h) = \beta\sigma_h$, with $\beta$ being a hand-picked scalar. Though they both apply to head positions, the use of $k$NN for $\sigma_h$ in the density map is not to be confused with the full $k$NN map used in our method, which is defined by,
\begin{multline}
    K(\bm{x}, k) =\\ \frac{1}{k}\sum\min\limits_{k}\left({\sqrt{(x-x_h)^2+(y-y_h)^2},  \forall \bm{h} \in \mathcal{H}}\right)\text{,}
\end{multline}
where $\mathcal{H}$ is the list of all head positions. In other words, the $k$NN distance from each pixel, $(x, y)$, to each head position, $(x_h, y_h)$, is calculated.

To produce the inverse $k$NN (i$k$NN) map, we use,
\begin{equation}
   M = \frac{1}{K(\bm{x}, k) + 1}\text{,}
\end{equation}
where $M$ is the resulting i$k$NN map, with the addition and inverse being applied element-wise.

\begin{figure}
    \centering
    \input{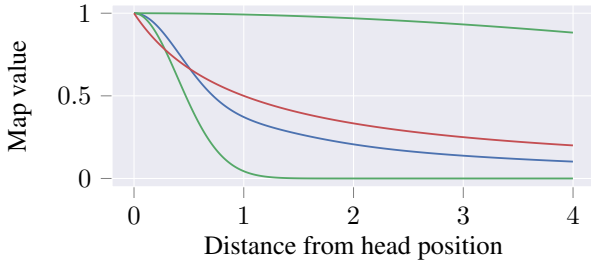}
    \caption{A comparison of the values of map labeling schemes with respect to the distance from an individual head position (normalized for comparison). Two Gaussians are shown in green. The narrow Gaussian provides a precise location of the head labeling. However, it provides little training information as the distance from the head increases. The wide Gaussian provides training information at a distance, but gives an imprecise location of the head position, resulting in low training information near the correct answer. The blue line shows a composite of several Gaussians with spread parameters between those of the two extremes (\cite{idrees2018composition} uses 3 Gaussian spreads in their work). This provides both precise and distant training losses. Our approach of the i$k$NN map shown in red (with $k=1$) approaches a map function with a shape similar to the integral on the spread parameter of all Gaussians for a spread parameter range from 0 to some constant. Additionally, our method provides both the precise and distant gradient training information in a single map label. Also notable, is that even the large Gaussian shown here approaches near zero much sooner than the i$k$NN map value.}
    \label{fig:mapcomparisons}
\end{figure}

To understand the advantage of an i$k$NN map over a density map, we can consider taking the generation of density maps to extremes with regard to the spread parameter of the Gaussian kernel provided by $f$. A similar explanation is illustrated in \cref{fig:mapcomparisons}. At one extreme, is a Gaussian kernel with zero spread. Here the delta function remains unchanged, which in practical terms translates to a density map where the density for each pedestrian is fully residing on a single pixel. When the difference between the true and predicted density maps is used to calculate a training loss, the network predicting density 1 pixel away from the correct labeling is considered just as incorrect as 10 pixels away from the correct labeling. Obviously, this is not desired, as it both creates a discontinuous training gradient, and the training process is intolerant to minor spatial labeling deviations. The other extreme is a very large Gaussian spread. This results in inexact spatial information of the location of the density. At the extreme, this provides no benefit over a global regression, which is the primary purpose for using a density map in the first place. Any intermediate Gaussian spread has an intermediate degree of both these problems. Using multiple scales of Gaussian spread, \cite{idrees2018composition} tries to obtain the advantage of both sides. However, the size of the scales and the number of scales are then arbitrary and hard to determine.

In contrast, a single i$k$NN map provides a substantial spatial gradient everywhere while still providing steep gradients in the exact locations of individual pedestrians. An example of our i$k$NN map compared with a corresponding density map labeling can be seen in \cref{fig:knnmap}. Notably, \cite{idrees2018composition} uses 3 density maps with different Gaussian spread parameters, with the Gaussian spread being determined by the $k$NN distance to other head positions multiplied by one of the 3 spread parameters. We note that for a single head position, all Gaussian distributions integrated over the spread parameter from 0 to some constant $\alpha$ results in a form of the incomplete gamma function. This function has a cusp around the center of the Gaussians. Similarly, the inverse of the $k$NN map also forms a cusp at the head position and results in similar gradients of loss given misplaced density/distance values as the spread integrated Gaussian function. In our experiments, we found that an inverse $k$NN map outperformed density maps with ideally selected spread parameters (and further outperformed general use selected spread parameters).

In one experiment, we use \cite{idrees2018composition}'s network architecture, which utilized DenseBlocks~\cite{huang2017densely} as the basis, but we replace the density maps with i$k$NN maps and show there is an improvement in the prediction's mean absolute error. This demonstrates the direct improvement of our i$k$NN method on an existing state-of-the-art network. Note, the regression module from i$k$NN map to count is then also required to convert from the i$k$NN map to a count. The difference in error between the original approach in \cite{idrees2018composition} and the network in \cite{idrees2018composition} with our i$k$NN maps, though improved, is relatively small. We suspect this is because the density maps (or i$k$NN maps) used during training are downsampled to a size of 28x28 (where the original images and corresponding labels are 224x224). This severe downsampling results in more binning of pixel information, and this seems to reduce the importance of which system is used to generate that label. At the extreme case, when downsampled to a single value, both approaches would only give the global count in the patch (where the i$k$NN map gives the inverse of the average distance from a pixel to a head labeling which can be translated to an approximate count). This downsampling is a consequence of the network structure only permitting labels of the same spatial size as the output of the DenseBlocks. Our network (which will be described below) remedies this through transposed convolutions, allowing for the use of the full-size labels.
    \section{MUD-i$k$NN: A New Network Architecture}
\label{sec:network}
\begin{figure*}
    \centering
    \includegraphics[width=0.85\textwidth]{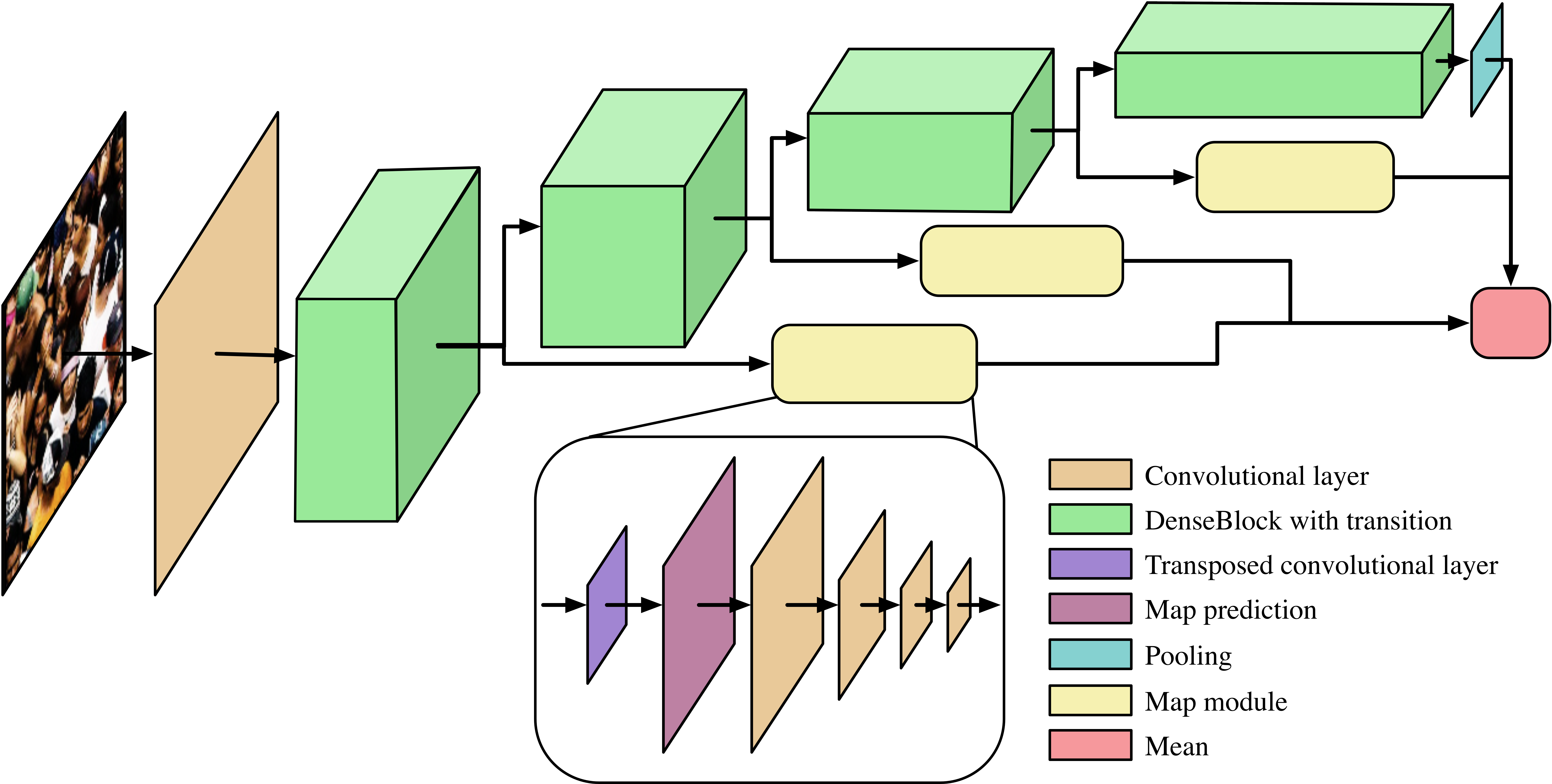}
    \caption{A diagram of the proposed network architecture MUD-i$k$NN: multiscale regression with DenseBlocks and i$k$NN mapping. Best viewed in color.}
    \label{fig:network}
\end{figure*}
We propose a new network structure, MUD-i$k$NN, with both multi-scale upsampling using DenseBlocks~\cite{huang2017densely} and our i$k$NN mapping scheme. We show that the new MUD-i$k$NN structure performs favorably compared with existing state-of-the-art networks. In addition to the use of i$k$NN maps playing a central role, we also demonstrate how features with any spatial size can contribute in the prediction of i$k$NN maps and counts through the use of transposed convolutions. This allows features of various scales from throughout the network to be used for the prediction of the crowd.

The proposed MUD-i$k$NN network structure is shown in  \cref{fig:network}. Our network uses the DenseBlock structures from DenseNet201~\cite{huang2017densely} in their entirety. DenseNet has been shown to be widely applicable to various problems. The output of each DenseBlock (plus transition layer) is used as the input to the following DenseBlock, just as it is in DenseNet201. However, each of these outputs is also passed to a transposed convolutional layer (excluding the final DenseBlock output). These transposed convolutions are given a stride and kernel size such that output is the size of the i$k$NN map, and no spatial input dimensions have their output overlap in the produced i$k$NN map. This form of upsampling allows the feature depth dimensions to contribute to the gradients of the map values in the predicted i$k$NN map. Both the stride and kernel size of the transposed convolutions of our network are 8, 16, and 32.

The i$k$NN map generated at each level is individually compared against the ground truth i$k$NN map, each producing a loss which is then summed,
\begin{equation}
    \mathcal{L}_{m} = \sum\limits_j\text{MSE}(\hat{M}_j, M_j)
\end{equation}
where $j$ is the index of the DenseBlock that the output came from, $M$ is the i$k$NN map labeling, and $\hat{M}$ is the predicted map labeling.

Each i$k$NN map is then also used as the input to a small regression module. This module is a series of small convolutional layers, shown in the inset of \cref{fig:network}. The sizes of these layers are specified in \cref{tab:mapmodule}. The regression module then has a singleton output, corresponding to the predicted crowd count.
\begin{table}
    \centering
    \begin{tabular}{c c c}  
        \toprule
        Layer & Output size & Filter \\
        \midrule
        \makecell{Input from\\DenseBlock} & \makecell{128x28x28\\256x14x14\\896x7x7} & \\
        \midrule
        \makecell{Transposed\\convolution} & \makecell{1x224x224\\(map prediction)} & \makecell{(8,16,32)x(8,16,32)\\stride=(8,16,32)} \\
        \midrule
        Convolution & 8x112x112 & 2x2 stride=2 \\
        \midrule
        Convolution & 16x56x56 & 2x2 stride=2 \\
        \midrule
        Convolution & 32x28x28 & 2x2 stride=2 \\
        \midrule
        Convolution & 1x1x1 & 28x28 \\
        \bottomrule
    \end{tabular}
    \caption{A specification of the map module layers. This module is used at 3 points throughout our network as shown in \cref{fig:network}, so the initial input size varies. However, the transposed convolution always produces a predicted map label which is uniform size (1x224x224).}
    \label{tab:mapmodule}
\end{table}

The mean of all predicted crowd counts from the regression modules, three in \cref{fig:network}, and the output of the final DenseBlock is used as the final count prediction.
\begin{equation}
\mathcal{L}_c = \text{MSE}\left( \frac{\hat{C}_{end} + \sum\limits_{j=1}^{m}\hat{C}_j}{m + 1}, C \right)
\end{equation}
with $C$ being the ground truth count, $\hat{C}_{end}$ being the regression count output by the final DenseBlock, and $\hat{C}_{j}$ being the count from the $j$th map regression module ($j=1,2,...,m; m=3$ in  \cref{fig:network}). This results in a total loss given by $\mathcal{L} = \mathcal{L}_{m} + \mathcal{L}_{c}$.

This approach has multiple benefits. First, if an appropriately sized stride and kernel size are specified, the transposed convolutional layer followed by i$k$NN map prediction to regression module can accept any sized input. This means this module of the network is very generalizable and can be applied to any CNN structure at any point in the network. For example, an additional DenseBlock could be added to either end of the DenseNet, and another of these map modules could be attached. Second, each i$k$NN map is individually trained to improve the prediction at that layer, which provides a form of intermediate supervision, easing the process of training earlier layers in the network. At the same time, the final count is based on the mean values of the regression modules. This means that if any individual regression module produces more accurate results, its results can individually be weighted as being more important to the final prediction.

We note that the multiple Gaussian approach by \cite{idrees2018composition} has some drawbacks. The spread of the Gaussians, as well as the number of different density maps, is somewhat arbitrary. Additionally, without upsampling, a separate network branch is required to maintain spatial resolution. This results in redundant network parameters and a final count predictor which is largely unconnected to the map prediction optimization goal. Our upsampling approach allows the main network to retain a single primary branch and connects all the optimization goals tightly to this branch.

The input to the network is 224$\times$224 image patches. The i$k$NN maps (or density maps) use the same size patches. Each map regression module contains the layers specified in \cref{tab:mapmodule}. At evaluation time, a sliding window with a step size of 128 was used for each patch of the test images, with overlapping predictions averaged.

Network code and hyperparameters can be found at \url{https://github.com/golmschenk/sr-gan}.

    \section{Experimental Results}
\label{sec:results}

\subsection{Evaluation metrics}
For each dataset that we evaluated our method on, we provide the mean absolute error (MAE), normalized absolute error (NAE), and root mean squared error (RMSE). These are given by the following equations:
\begin{equation}
    \text{MAE} = \frac{1}{N}\sum\limits_{i=1}^{N}\abs{\hat{C}_i - C_i}
\end{equation}
\begin{equation}
\text{NAE} = \frac{1}{N}\sum\limits_{i=1}^{N}\frac{\abs{\hat{C}_i - C_i}}{C_i}
\end{equation}
\begin{equation}
\text{RMSE} = \sqrt{\frac{1}{N}\sum\limits_{i=1}^{N}(\hat{C}_i - C_i)^2}
\end{equation}

In the first set of experiments, we demonstrate the improvement of the i$k$NN labeling scheme compared to the density labeling scheme. We trained our network using various density maps produced with different Gaussian spread parameters, $\beta$ (as described in \cref{sec:knnmaps}), and compared these results to the network using i$k$NN maps with varying $k$. We also analyze the advantage of upsampling the label for both density and i$k$NN maps. In the second set of experiments, we provide comparisons to the state-of-the-art on standard crowd counting datasets. In these comparisons, the best i$k$NN map and density map from the first set of experiments is used. Most works provide their MAE and RMSE results. \cite{idrees2018composition} provided the additional metric of NAE. Though this result is not available for many of the datasets, we provide our own NAE on these datasets for future works to refer to. The most directly relevant work, \cite{idrees2018composition}, has only provided their results for their latest dataset, UCF-QNRF. As such, their results only appear in regard to that dataset. Finally, we offer a general analysis of the results using our i$k$NN maps and upsampling approaches. General statistics about the datasets used in our experiments is shown in \cref{tab:Dataset statistics}.

\subsection{Impact of labeling approach and upsampling}

\subsubsection{Density maps vs i$k$NN maps}

We used the ShanghaiTech dataset~\cite{zhang2016single} part A for this analysis. The results of these tests are shown in \cref{tab:dishanghaitechparta}. The density maps provide a curve, where too large and too small of spreads perform worse than an intermediate value. Even when choosing the best value (where $\beta = 0.3$), which needs to manually determined, the i$1$NN label significantly outperforms the density label.

\begin{table*}
    \centering
    \begin{tabular}{lccccc}  
        \toprule
        Dataset & Images & Total count  & Mean count & Max count & Average resolution\\
        \midrule
        UCF-QNRF & 1535 & 1,251,642 & 815 & 12,865 & 2013$\times$2902\\
        ShanghaiTech Part A & 482 & 241,677 & 501 & 3139 & 589$\times$868 \\
        ShanghaiTech Part B & 716 & 88,488 & 123.6 & 578 & 768$\times$1024 \\
        UCF-CC-50 & 50 & 63,974 & 1279 & 4633 & 2101$\times$2888 \\
        \bottomrule
    \end{tabular}
    \caption{General statistics for the tested datasets.}
    \label{tab:Dataset statistics}
\end{table*}
\begin{table}
    \centering
    \begin{tabular}{lccc}  
        \toprule
        Method & MAE & NAE & RMSE \\
        \midrule
        \textbf{MUD-density$\beta0.3$} 28x28 & 79.0 & 0.209 & 120.5 \\
        \textbf{MUD-density$\beta0.3$} 56x56 & 74.8 & 0.181 & 121.0 \\
        \textbf{MUD-density$\beta0.3$} 112x112 & 73.3 & 0.176 & 119.1 \\
        \textbf{MUD-i$1$NN} 28x28 & 75.8 & 0.180 & 120.3 \\
        \textbf{MUD-i$1$NN} 56x56 & 72.7 & 0.181 & 117.4 \\
        \textbf{MUD-i$1$NN} 112x112 & 70.8 & 0.166 & 117.0 \\
        \midrule
        \textbf{MUD-density$\beta0.05$} & 84.5 & 0.233 & 139.9 \\
        \textbf{MUD-density$\beta0.1$} & 76.8 & 0.189 & 120.3 \\
        \textbf{MUD-density$\beta0.2$} & 75.3 & 0.175 & 124.2 \\
        \textbf{MUD-density$\beta0.3$} & 72.7 & 0.174 & 120.4 \\
        \textbf{MUD-density$\beta0.4$} & 75.7 & 0.176 & 130.5 \\
        \textbf{MUD-density$\beta0.5$} & 76.3 & 0.182 & 130.0 \\
        \textbf{\makecell[l]{MUD-density\\\quad$\beta_{1}0.5$,$\beta_{2}0.3$,$\beta_{3}0$}} & 78.5 & 0.205 & 124.2 \\
        \textbf{\makecell[l]{MUD-density\\\quad$\beta_{1}0.5$,$\beta_{2}0.3$,$\beta_{3}0.05$}} & 77.8 & 0.207 & 124.9 \\
        \textbf{\makecell[l]{MUD-density\\\quad$\beta_{1}0.4$,$\beta_{2}0.2$,$\beta_{3}0.1$}} & 76.7 & 0.202 & 122.7 \\
        \textbf{\makecell[l]{MUD-density\\\quad$\beta_{1}0.1$,$\beta_{2}0.2$,$\beta_{3}0.4$}} & 75.1 & 0.191 & 119.0 \\
        \textbf{\makecell[l]{MUD-density\\\quad$\beta_{1}0.2$,$\beta_{2}0.3$,$\beta_{3}0.4$}} & 76.0 & 0.196 & 122.1 \\
        \midrule
        \textbf{MUD-i$1$NN} & 68.0 & 0.162 & 117.7 \\
        \textbf{MUD-i$2$NN} & 68.8 & 0.168 & 109.0 \\
        \textbf{MUD-i$3$NN} & 69.8 & 0.169 & 110.7 \\
        \textbf{MUD-i$4$NN} & 72.2 & 0.173 & 116.0 \\
        \textbf{MUD-i$5$NN} & 74.0 & 0.182 & 119.1 \\
        \textbf{MUD-i$6$NN} & 76.2 & 0.188 & 120.9 \\
        \bottomrule
    \end{tabular}
    \caption{Results using density maps vs i$k$NN maps with varying $k$ and $\beta$, as well as the various upsampling resolutions on the ShanghaiTech Part A dataset. If a resolution is not shown, it is the default 224$\times$224. Multiple $\beta$ correspond to a different Gaussian density map for each of the 3 map module comparisons.}
    \label{tab:dishanghaitechparta}
\end{table}

Included in the table are experiments, in the fashion of \cite{idrees2018composition}, with density maps using 3 different $\beta$ values. Here $\beta_1$ denotes the spread parameter used as the label map for the first map module, while $\beta_2$ and $\beta_3$ are for the second and third modules. Contrary to \cite{idrees2018composition}'s findings, we only gained a benefit from 3 density labels when the first output had the smallest spread parameter. Even then, the gain was minimal. Upon inspection of the weights produced by the network from the map to the count prediction, the network reduces the predictions from the non-optimal $\beta$ maps to near zero and relies solely on the optimal map (resulting in a reduced accuracy compared to using the optimal map for each map module).

With varying $k$, we find that an increased $k$ results in lower accuracy. This is likely due to the loss of precision in the location of an individual. The most direct explanation for this can be seen in the case of $k=2$. Every pixel on the line between two nearest head positions will have the same map value, thus losing the precision of an individual location.

\subsubsection{Upsampling analysis}
\begin{figure*}
    \centering
    \begin{subfigure}{0.5\textwidth}
      \centering
      \includegraphics[width=0.98\textwidth,height=0.55\textwidth]{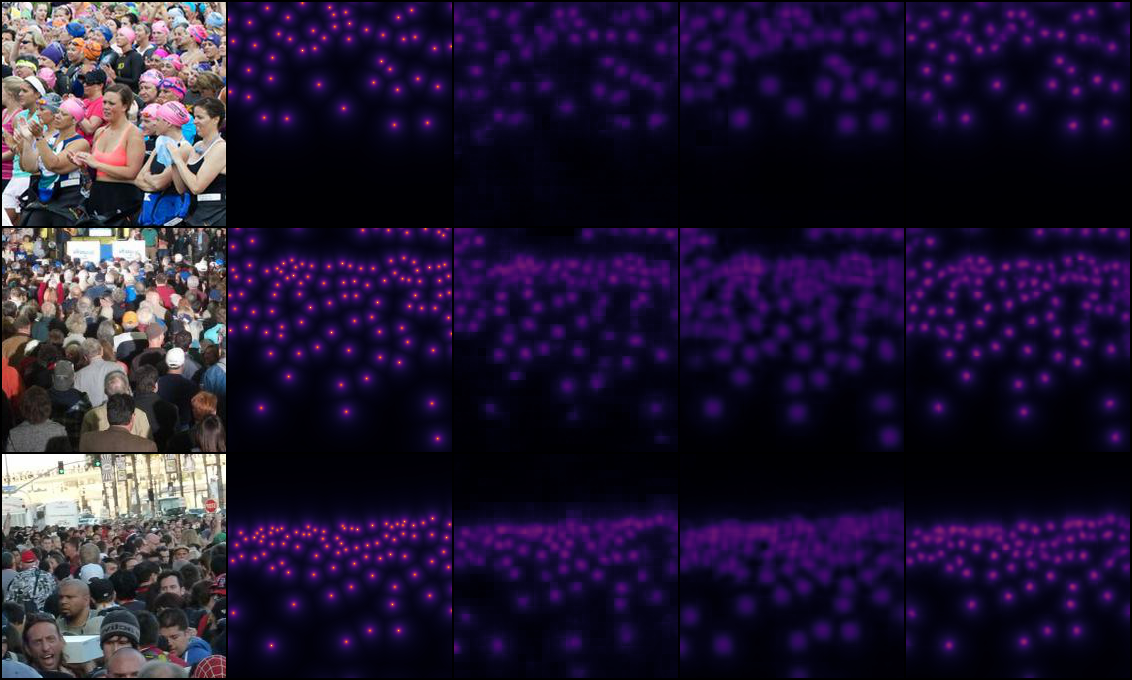}
      \caption{i$1$NN predictions.}
      \label{fig:sub1}
    \end{subfigure}%
    \begin{subfigure}{0.5\textwidth}
      \centering
      \includegraphics[width=0.98\textwidth,height=0.55\textwidth]{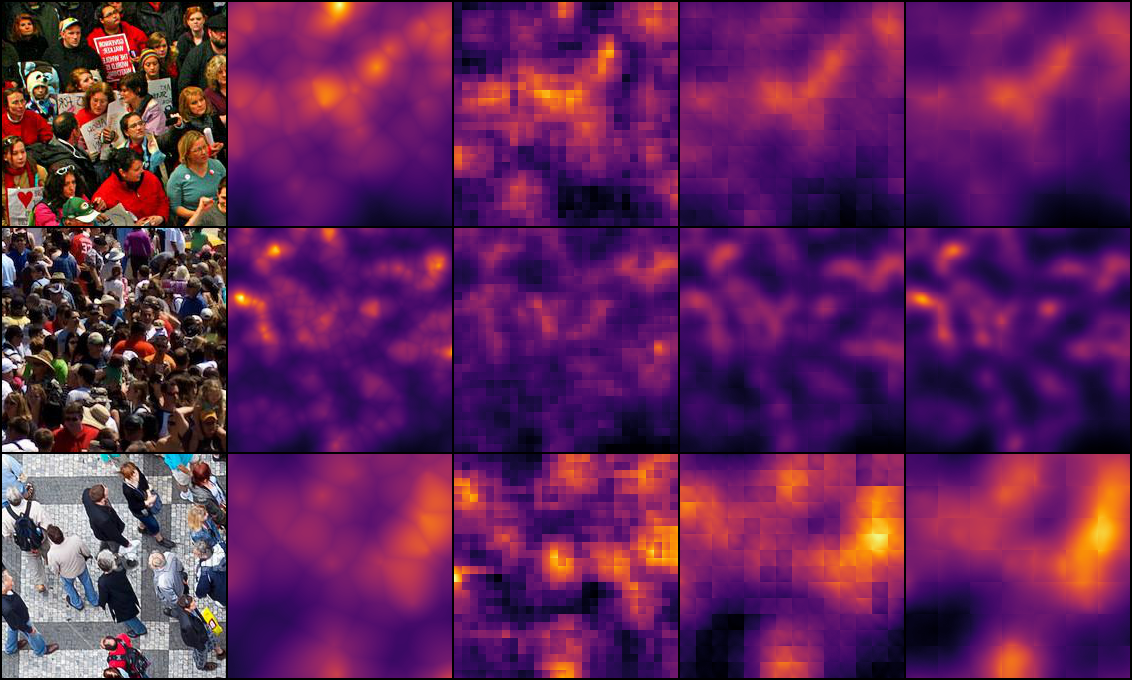}
      \caption{i$3$NN predictions.}
      \label{fig:sub2}
    \end{subfigure}
    \caption{A small sample of patch predictions for map labels. In each subfigure, from left to right is the original image patch, the ground truth label, and the patches from the three map modules in order through the network.}
    \label{fig:predictedmaps}
\end{figure*}
Most existing works use a density map with a reduced size label for testing and training. Those that use the full label resolution design specific network architectures for the high-resolution labels. Our map module avoids this constraint by upsampling the label using a trained transposed convolution, which can be integrated into most existing architectures. Using the ShanghaiTech part A dataset, we tested our network using various label resolutions to determine the impact on the predictive abilities of the network. These results can be seen in \cref{tab:dishanghaitechparta}. Experiments without no label resolution given are 224$\times$224. From these results, it is clear that the higher resolution leads to higher accuracy. Note, this results in a minor change to the map module structure, as the final convolution kernel needs to match the remaining spatial dimension. A set of predicted i$k$NN map labels can be seen in \cref{fig:predictedmaps}, where a grid pattern due to the upsampling can be identified in some cases.

\subsection{Comparisons on standard datasets}
The following demonstrates our network's predictive capabilities on various datasets, compared to various state-of-the-art methods. Again, we note that our improvements are expected to complementary to the existing approaches, rather than alternatives.

For these experiments, we used the best $k$, 1, and best $\beta$, 0.3, from the first set of experiments.

The first dataset we evaluated our approach on is the UCF-QNRF dataset~\cite{idrees2018composition}. The results of our MUD-i$k$NN network compared with other state-of-the-art networks are shown in \cref{tab:ufc-qnrf-results}. Our network significantly outperforms the existing methods. Along with a comparison of our complete method compared with the state-of-the-art, we compare with \cite{idrees2018composition}'s network, but replace their density map predictions and summing to count with our i$k$NN map prediction and regression to count. Using the i$k$NN maps, we see that their model sees improvement in MAE with i$k$NN maps, showing the effect of the i$k$NN mapping.

\begin{table}
    \centering
    \begin{tabular}{lccc}  
        \toprule
        Method & MAE & NAE & RMSE \\
        \midrule
        Idrees \etal(2013)~\cite{idrees2013multi} & 315 & 0.63 & 508 \\
        MCNN~\cite{zhang2016single} & 277 & 0.55 & 426 \\
        Encoder-Decoder~\cite{badrinarayanan1511deep} & 270 & 0.56 & 478 \\
        CMTL~\cite{sindagi2017cnn} & 252 & 0.54 & 514 \\
        SwitchCNN~\cite{sam2017switching} & 228 & 0.44 & 445 \\
        Resnet101~\cite{he2016deep} & 190 & 0.50 & 227 \\
        DenseNet201~\cite{huang2017densely}& 163 & 0.40 & 226 \\
        Idrees \etal(2018)~\cite{idrees2018composition} & 132 & 0.26 & 191 \\
        \midrule
        \textbf{\cite{idrees2018composition} with i$1$NN maps} & 122 & 0.252 & 195 \\
        \textbf{MUD-i$1$NN} & 104 & 0.209 & 172 \\
        \bottomrule
    \end{tabular}
    \caption{Results on the UCF-QNRF dataset.}
    \label{tab:ufc-qnrf-results}
\end{table}

The second dataset we evaluated our approach on is the ShanghaiTech dataset~\cite{zhang2016single}. The dataset is split into two parts, Part A and Part B. For both parts, we used the training and testing images as prescribed by the dataset provider. The results of our evaluation on part A are shown in \cref{tab:shanghaitechparta}. Our MUD-i$k$NN network slightly outperforms the state-of-the-art approaches on this part. The results of our evaluation on part B are shown in \cref{tab:shanghaitechpartb}. Here our network performs on par or slightly worse than the best-performing methods.

\begin{table}
    \centering
    \begin{tabular}{lccc}  
        \toprule
        Method & MAE & NAE & RMSE \\
        \midrule
        ACSCP~\cite{shen2018crowd} & 75.7 & - & 102.7 \\
        D-ConvNet-v1\cite{shi2018crowd} & 73.5 & - & 112.3 \\
        ic-CNN~\cite{ranjan2018iterative} & 68.5 & - & 116.2 \\
        CSRNet~\cite{li2018csrnet} & 68.2 & - & 115.0 \\
        \midrule
        \textbf{MUD-density$\beta0.3$} & 72.7 & 0.174 & 120.4 \\
        \textbf{MUD-i$1$NN} & 68.0 & 0.162 & 117.7 \\
        \bottomrule
    \end{tabular}
    \caption{Results on the ShanghaiTech Part A dataset.}
    \label{tab:shanghaitechparta}
\end{table}
\begin{table}
    \centering
    \begin{tabular}{lccc}  
        \toprule
        Method & MAE & NAE & RMSE \\
        \midrule
        D-ConvNet-v1\cite{shi2018crowd} & 18.7 & - & 26.0 \\
        ACSCP~\cite{shen2018crowd} & 17.2 & - & 27.4 \\
        ic-CNN~\cite{ranjan2018iterative} & 10.7 & - & 16.0 \\
        CSRNet~\cite{li2018csrnet} & 10.6 & - & 16.0 \\
        \midrule
        \textbf{MUD-density$\beta0.3$} & 16.6 & 0.130 & 26.9 \\
        \textbf{MUD-i$1$NN} & 13.4 & 0.107 & 21.4 \\
        \bottomrule
    \end{tabular}
    \caption{Results on the ShanghaiTech Part B dataset.}
    \label{tab:shanghaitechpartb}
\end{table}

The third dataset we evaluated our approach on is the UCF-CC-50 dataset~\cite{idrees2013multi}. We followed the standard evaluation metric for this dataset of a five-fold cross-evaluation. The results of our evaluation on this dataset can be seen in \cref{tab:ucfcc50}.

\begin{table}
    \centering
    \begin{tabular}{lccc}  
        \toprule
        Method & MAE & NAE & RMSE \\
        \midrule
        ACSCP~\cite{shen2018crowd} & 291.0 & - & 404.6 \\
        D-ConvNet-v1\cite{shi2018crowd} & 288.4 & - & 404.7 \\
        CSRNet~\cite{li2018csrnet} & 266.1 & - & 397.5 \\
        ic-CNN~\cite{ranjan2018iterative} & 260.9 & - & 365.5 \\
        \midrule
        \textbf{MUD-density$\beta0.3$} & 246.44 & 0.188 & 348.1 \\
        \textbf{MUD-i$1$NN} & 237.76 & 0.191 & 305.7 \\
        \bottomrule
    \end{tabular}
    \caption{Results on the UCF-CC-50 dataset.}
    \label{tab:ucfcc50}
\end{table}

Overall, our network performed favorably compared with existing approaches. An advantage to our approach is that the our modifications can be applied to the architectures we're comparing against. The most relevant comparison is between the i$k$NN version of the MUD network, and the density map version. Here, the i$k$NN approach always outperformed the density version. We speculate that the state-of-the-art methods we have compared with, along with other general-purpose CNNs, could be improved through the use of i$k$NN labels and upsampling map modules.
    \section{Conclusions}
\label{sec:conclusions}
In this work, we have presented a new form of labeling for crowd counting data, the i$k$NN map. We have compared this labeling scheme to commonly accepted labeling approach for crowd counting, the density map. We show that using the i$k$NN map with an existing state-of-the-art network improves the accuracy of the network compared to density map labelings. We have demonstrated the improvements gained by using increased label resolutions, and provide an upsampling map module which can be generally used by other crowd counting architectures. These approaches can easily be incorporated into other crowd counting techniques, as we have incorporated them into DenseNet, which resulted in a network which performs favorably compared with the state-of-the-art.

    {\small
    \bibliographystyle{ieee}
    \bibliography{bibliography}
    }

\end{document}